\documentclass{article}

\PassOptionsToPackage{numbers, compress}{natbib}

\usepackage[preprint]{neurips_2022}

\usepackage[utf8]{inputenc} 
\usepackage[T1]{fontenc}    
\usepackage{hyperref}       
\usepackage{url}            
\usepackage{booktabs}       
\usepackage{graphicx}
\usepackage{amsmath,amsfonts}       
\usepackage{nicefrac}       
\usepackage{microtype}      
\usepackage{multirow}
\usepackage[dvipsnames,table,xcdraw]{xcolor}

\hypersetup{
    colorlinks=true,
    linkcolor=BrickRed,
    citecolor=RoyalPurple,
    filecolor=magenta,      
    urlcolor=RoyalBlue,
}

\newlength\savewidth\newcommand\shline{\noalign{\global\savewidth\arrayrulewidth
		\global\arrayrulewidth 1pt}\hline\noalign{\global\arrayrulewidth\savewidth}}
\newcommand{\tablestyle}[2]{\setlength{\tabcolsep}{#1}\renewcommand{\arraystretch}{#2}\centering}

\def\ie{\emph{i.e.,~}}
\def\eg{\emph{e.g.,~}}
\def\etal{{\em et al.~}}
\def\etc{{\em etc.~}}
\def\wrt{\emph{w.r.t.~}}

\newcommand{\figref}[1]{Figure~\ref{#1}}
\newcommand{\tabref}[1]{Table~\ref{#1}}
\newcommand{\secref}[1]{Section~\ref{#1}}

\usepackage[misc]{ifsym}
\newcommand\blfootnote[1]{%
\begingroup
\renewcommand\thefootnote{}\footnote{#1}%
\addtocounter{footnote}{-1}%
\endgroup
}

\title{Delving Deeper into Data Scaling \\in Masked Image Modeling}

\author{%
  Cheng-Ze Lu$^{1*}$ \quad Xiaojie Jin$^{2\,}$\textsuperscript{\Letter} \quad Qibin Hou$^{1}$ \quad Jun Hao Liew$^{2}$ \\ 
  \textbf{Ming-Ming Cheng$^{1}$ Jiashi Feng$^{2\,}$\textsuperscript{\Letter}} \\
  $^{1}$ VCIP, CS, Nankai University. \qquad
  $^{2}$ ByteDance Inc.
}

\begin{document}

\blfootnote{$^{*}$ Work done at Bytedance Inc.  \newline \hspace*{1.9em}{\scriptsize \Letter} Corresponding author: Xiaojie Jin$<$\url{jinxiaojie@bytedance.com}$>$, Jiashi Feng$<$\url{jshfeng@byte} \url{dance.com}$>$}
\maketitle

\begin{abstract}
Understanding whether self-supervised learning methods can scale with unlimited data is crucial for training large-scale models.
In this work, we conduct an empirical study on the scaling capability of masked image modeling (MIM) methods (\textit{e.g.,} MAE) for visual recognition.
Unlike most previous works that depend on the widely-used ImageNet dataset, which is manually curated and object-centric, we take a step further and propose to investigate this problem in a more practical setting. Specifically, we utilize the web-collected Coyo-700M dataset.
We randomly sample varying numbers of training images from the Coyo dataset and construct a series of sub-datasets, containing 0.5M, 1M, 5M, 10M, and 100M images, for pre-training.
Our goal is to investigate how the performance changes on downstream tasks when scaling with different sizes of data and models. 
The study reveals that:
1) MIM can be viewed as an effective method to improve the model capacity when the scale of the training data is relatively small;
2) Strong reconstruction targets can endow the models with increased capacities on downstream tasks;
3) MIM pre-training is data-agnostic under most scenarios, which means that the strategy of sampling pre-training data is non-critical.
We hope these observations could provide valuable insights for future research on MIM.

\end{abstract}

\section{Introduction}
Self-supervised pre-training has been proven an effective way to improve the model capacity for visual tasks.
Currently, two prominent methods dominate self-supervised pre-training: contrastive learning and masked image modeling (MIM).
Contrastive learning-based techniques, exemplified by SimCLR~\cite{chen2020simple} and MOCO~\cite{he2020momentum}, focus on minimizing the distance between representations of different views of the same image while maximizing the distance between representations of views from different images.
This type of methods naturally endows the pre-trained models with strong instance discriminability.
On the other hand, MIM-based methods, represented by MAE~\cite{he2022masked}, SimMIM~\cite{xie2022simmim}, and BEiT~\cite{bao2021beit}, aim to capture knowledge about local relationships within an input image through a reconstruction task.
Consequently, these methods opt to learn more expressive feature representations, enabling the pre-trained models to achieve remark performance in downstream tasks, like object detection and semantic segmentation.

Scalability is a crucial aspect of both supervised and unsupervised learning paradigms and can be examined from two perspectives: model scalability and data scalability.
In the field of Natural Language Processing, self-supervised masked language modeling has established the scaling law~\cite{kaplan2020scaling} and successfully trained large-scale language models~\cite{devlin2018bert,jia2021scaling}.
However, in the context of masked image modeling, while it has been proposed to support scalability in terms of model size, the question of data scalability remains unanswered.
%
%
Recent work~\cite{xie2022data} argues that MIM-based methods are scalable learners and still demanding for larger data when we use longer training lengths.
We propose to think about the problem of data scaling from a different perspective and provide different insights.

The development of large-scale multi-modal models in recent years has granted us access to web-scale datasets.
As a result, instead of using the ImageNet-1k dataset~\cite{deng2009imagenet}, which is manually processed and object-centric, for MIM pre-training, we select a larger yet more diverse dataset, Coyo-700M~\cite{kakaobrain2022coyo-700m}, to systematically study the data scalability of MIM pre-training.
In our experiments, we adopt the MAE~\cite{he2022masked} configurations, with the exception of the reconstruction target, which is produced by a target encoder. 
Under these conditions, we carefully observe and draw the following conclusions:
\begin{itemize}
    \item Data scaling is limited to 10M under the same model size for MIM pre-training. Over 10M data, pre-training will probably lead to performance saturation on most tasks, with the exception of the long-tailed LVIS~\cite{gupta2019lvis} detection where performances improves as the dataset size increases.
    \item A strong target encoder can endow the model with relatively better performance. However, it cannot break the limit of performance saturation.
    \item MIM pre-training could be data agnostic. Sampling from web-scale datasets with different strategies may not help with respect to downstream performances.
\end{itemize}
These findings are expected to provide valuable insights and contribute to the research community. 
\begin{figure}[t]
\centering
\includegraphics[width=0.97\textwidth]{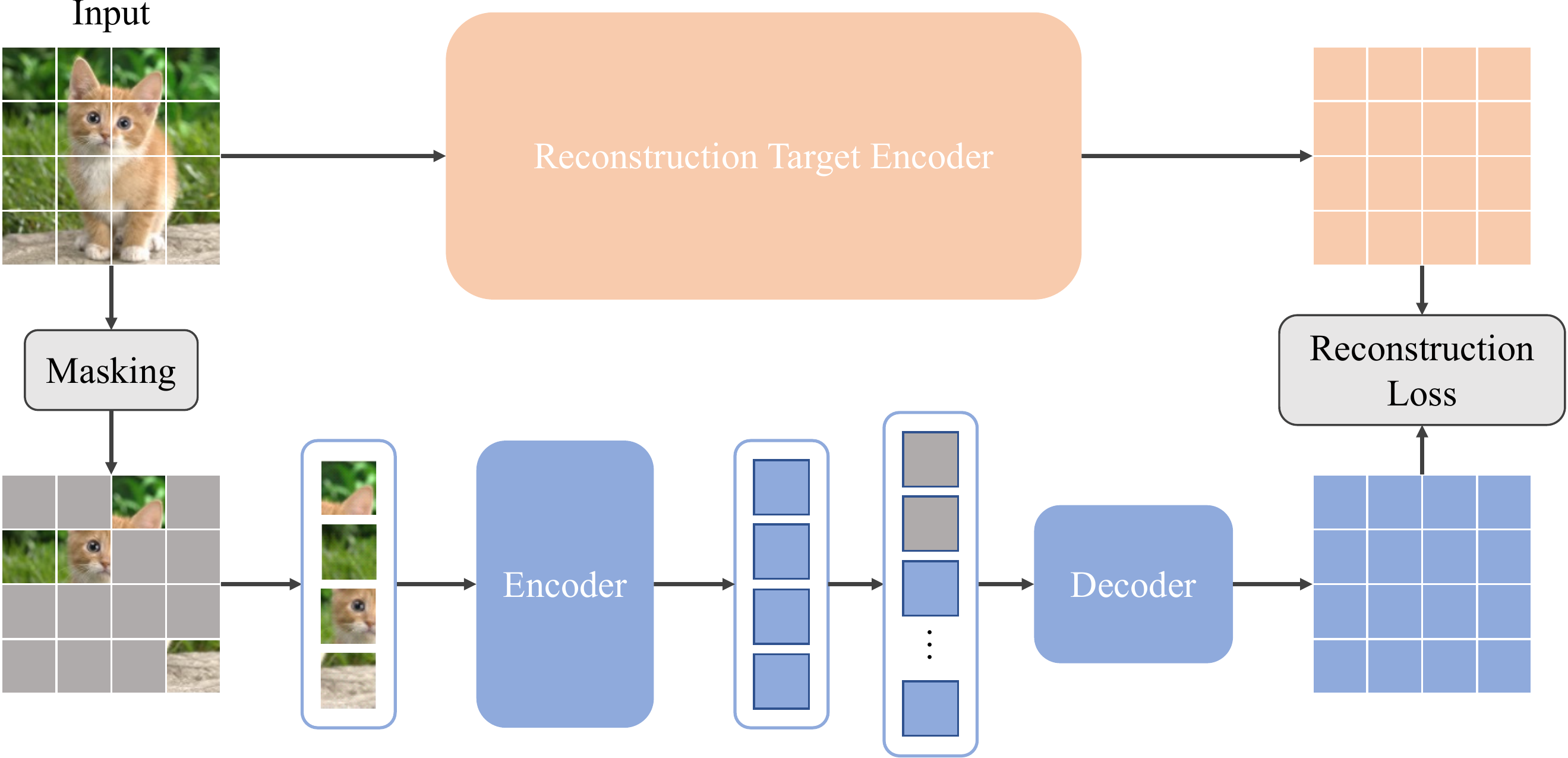}\label{fig:overall_framework}
\caption{The framework used for investigating the data scaling problem in this work.}
\end{figure}

\section{Related Work}
\paragraph{Masked Image Modeling}
Self-supervised learning, which focuses on acquiring powerful and transferable representations by leveraging data without human-labeled annotations, has garnered increasing attention~\cite{bao2021beit,he2022masked,huang2022contrastive,gao2022convmae,xie2022simmim,tao2022siamese,chen2022context} in recent years.
Recently, inspired by the success of masked language modeling in Natural Language Processing, masked image modeling methods like BEiT~\cite{bao2021beit}, SimMIM~\cite{xie2022simmim}, and MAE~\cite{he2022masked} have shown remarkable capabilities of representation in a ``mask-and-predict'' manner.
With a large portion of patches masked, these models directly predict the missed pixels~\cite{bao2021beit,he2022masked} or discrete visual vocabularies~\cite{bao2021beit}. 
Subsequent works try to speed up the convergence and import the performance by substituting the reconstruction targets for injecting semantic information~\cite{wei2022mvp,peng2022unified,hou2022milan,liu2022devil}, combining contrastive learning for strengthening the discriminability~\cite{huang2022contrastive,tao2022siamese}, or modifying the architecture for introducing locality~\cite{gao2022convmae}.
In this work, we tend to evaluate the data scaling ability under MAE~\cite{he2022masked}, which is the simplest framework of masked image modeling, with only the reconstruction target changed.

\paragraph{Scale-up Vision Models}
How to scale up models is critical in the recent large-model deep learning era.
Previously, based on MobileNets~\cite{howard2017mobilenets}, EfficientNet~\cite{tan2019efficientnet} proposes a scaling strategy and achieves a trade-off among depth, width, and resolution.
Zhai~\etal~\cite{zhai2022scaling} study the scaling law of Vision Transformers~\cite{dosovitskiy2020image} and try to model the relationship between the performance, data, and computation.
They successfully scale up a ViT model to 2 billion parameters with JFT-3B dataset.
Dehghani~\etal~\cite{dehghani2023scaling} propose a 22B-parameter ViT trained with an efficient and stable recipe and show the strong scaling potential of ViTs.

Many works~\cite{zhai2022scaling,alabdulmohsin2022revisiting} on scaling laws have been explored in supervised learning methods, while for self-supervised methods the scaling law still remains uncovered.
Masked image modeling is proven to be a scalable method~\cite{he2022masked} that as the model size scales up, the performance improves considerably.
EVA~\cite{fang2022eva} successfully train a 1B ViT model only using 30M pre-training data under MIM pre-training.
However, few works pay attention to data scaling of self-supervised learning methods.
Xie~\etal~\cite{xie2022data} study how the performances change when the dataset scales up, and observe that training length matters for pre-training.
Due to the object-centric property of ImageNet-1k dataset, we study data scaling on more natural datasets and find that the data scaling ability of masked image modeling is limited.

\section{Method and Experimental Setup}
\subsection{Method}
The overall framework of our method is shown in~\figref{fig:overall_framework}.
We follow the design of MAE~\cite{he2022masked}, except for the reconstruction target.
The input image $I$ is first split and projected into tokens $\{x_i\}_{i=1}^N$, with $N$ being the length of the token sequence.
Position embedding is then added.
After random masking, only visible tokens are fed into the encoder. Following MAE, we adopt Vision Transformer (ViT)~\cite{dosovitskiy2020image} architecture as the encoder.
Before feeding to the decoder, mask tokens and visible tokens are put together and rearranged according to its original order.
Here, position embedding is added for encoding location information.
The decoder part consists of another series of Transformer blocks to perform the reconstruction task, where the reconstruction target is produced by the output of the target encoder with access to the full set of input image tokens.
The reconstruction loss can be written as follows:
\begin{align}
    \mathcal{L}_r = \frac{(\overline{y}^t_{cls} - \overline{y}^s_{cls})^2 + \sum_{i=1}^{N}(\overline{y}^t_i - \overline{y}^s_i)^2}{N + 1},
\end{align}
where $\overline{y}^t$ and $\overline{y}^s$ denote the $l_2$-normalized output of the target encoder and the decoder.
Note that we include the [CLS] token with both masked and unmasked patches to calculate the reconstruction loss.
Different from MAE which uses pixel values as the reconstruction target, many recent works~\cite{wei2022mvp,fang2022eva,hou2022milan} replace the RGB target with a language-assisted target, and show that the reconstruction target actually matters since the reconstruction target determines what semantics to be learned by the encoder.
Here, we use a target encoder to produce the reconstruction target for further investigation.

\subsection{Architecture Setup}
As for the encoder, we evaluate three different ViT variants on downstream tasks, including ViT-B/16, ViT-L/16, and ViT-H/16, whose parameters number ranges from \textasciitilde 90M to \textasciitilde 650M.
The decoder consists of a stack of Transformer blocks (4 blocks by default). 
We adopt ViT-B/16 as the architecture of the target encoder for feature alignment. In this work, we investigate various target encoders, including MAE~\cite{he2022masked}, DINO~\cite{caron2021emerging}, CMAE~\cite{huang2022contrastive}, CLIP~\cite{radford2021learning},~\etc

\subsection{Pre-training Datasets}
We first pre-train ViT-B/16 on ImageNet-1k and ImageNet-22k datasets to study the effect of different target encoders.
Considering that the images in the ImageNet dataset are object-centric and fail to represent the complexity and inter-object relationships found in realistic natural scenes,
we thus choose the more diverse Coyo-700M dataset~\cite{kakaobrain2022coyo-700m} to study the problem of data scalability.
We randomly sample images from the Coyo-700m dataset to form 5 sub-datasets, namely Coyo-0.5m, Coyo-1m, Coyo-5m, Coyo-10m, and Coyo-100m. Furthermore, for a fair comparison, each small dataset is a subset of the larger one.
In addition, to examine the influence of different data sampling strategies, we adopt CiT method~\cite{xu2023cit} as the data sampling strategy to obtain higher-quality data.
Detailed experimental settings can be found in the supplementary.

\subsection{Pre-training Details}
When exploring the impact of different reconstruction objectives on data scalability, we pre-train models on ImageNet-1k~\cite{deng2009imagenet} and ImageNet-22k~\cite{russakovsky2015imagenet} datasets for 300 epochs and 90 epochs with 40 epochs for warming up, respectively. 
When experimenting with more diverse and natural images, \ie Coyo-\{0.5m, 1m, 5m, 10m, 100m\}, we conduct pre-training with \{300, 300, 300, 90, 15\} epochs, respectively.
The batch size is always set as 4096 during pre-training, and the masking ratio is set as 75\% following~\cite{he2022masked}.
We use AdamW optimizer ($base\_lr$=1.5e-4, $\beta_1, \beta_2$=0.9, 0.95, $weight\_decay$=0.05) with the cosine learning rate decay strategy.
We use the same augmentation strategy as MAE~\cite{he2022masked}, including random resize cropping and random flipping.
Detailed configurations and hyper-parameters are provided in the supplementary.

\subsection{Downstream Task Details}
We evaluate the pre-trained models on various downstream tasks, categorized as follows: (1) recognition tasks, including fine-tuning and linear-probing on ImageNet-1k~\cite{deng2009imagenet}, fine-tuning on iNaturalist-2018~\cite{van2018inaturalist}; (2) object detection and instance segmentation tasks on Microsoft COCO~\cite{lin2014microsoft} and LVISv1.0~\cite{gupta2019lvis}; (3) semantic segmentation tasks on ADE20K~\cite{zhou2017scene} and CityScapes~\cite{cordts2016cityscapes}.
To evaluate whether a large-scale pre-training dataset matters when dealing with ``harder'' downstream tasks, we select 5000 classes with the most number of images, namely ImageNet-5k, from ImageNet-22k and randomly split them into train and validation set for fine-tuning.
ImageNet-5k contains \textasciitilde6M and 0.25M images for training and validation, respectively.

\paragraph{ImageNet-1/5k and iNaturalist-2018.}
We conduct fine-tuning and linear-probing on ImageNet-1k dataset, which is most frequently used for image classification task.
As for fine-tuning, most configurations we use are the same as~\cite{he2022masked}, and detailed settings can be found in the supplementary.
We fine-tune ViT-\{B, L, H\}/16 for \{100, 50, 50\} epochs with 5 epochs of warming up.
AdamW optimizer~\cite{loshchilov2017decoupled} is adopted, and the base learning rate is set as 1e-3.
Layer-wise learning rate decay is also used and set as \{0.65, 0.75, 0.75\} for ViT-\{B, L, H\}/16.
On ImageNet-5k, we use the same strategy and hyper-parameters for fine-tuning to evaluate the performance of pre-trained model on this harder downstream task.
Linear-probing is another popular metric for evaluating the quality of representation.
We follow the configurations and hyper-parameter settings of MAE, and train ViT-\{B, L, H\}/16 for \{90, 50, 50\} epochs, and the warm-up epoch is set as 10.
Additionally, we follow the above experimental settings and fine-tune on iNaturalist-2018 dataset, which contains long-tailed fine-grained categories, for evaluating the learned representation on the long-tailed classification task.

\paragraph{Microsoft COCO and LVISv1.0.}
We follow~\cite{he2022masked}, and use Mask R-CNN~\cite{he2017mask} as the detector on COCO~\cite{lin2014microsoft} and LVIS~\cite{gupta2019lvis} dataset. 
Similar to ViTDet~\cite{li2022exploring}, we adapt ViT backbone with FPN~\cite{lin2017feature} and use random flipping, large-scale jittering with a scale ranging from 0.1 to 2.0, and random cropping for data augmentation on COCO dataset.
The batch size is set to 64, and trained on 64 GPUs.
AdamW optimizer ($lr$=1e-4, $\beta_1, \beta_2$=0.9, 0.999) is used with step-wise learning rate decay on COCO dataset.
We report $AP^{box}$ for object detection and $AP^{mask}$ for instance segmentation.
LVIS is a long-tailed large-vocabulary dataset with more than 1,200 categories for instance segmentation. As a result, LVIS dataset is more challenging than COCO.
Apart from the above augmentation, we also employ repeat factor sampling strategy~\cite{gupta2019lvis} to deal with long-tailed classes.
We adopt AdamW optimizer ($lr$=2e-4 for ViT-B/L, 1e-4 for ViT-H, $\beta_1, \beta_2$=0.9, 0.999) on LVIS dataset.
Additionally, we also report $AP^{r}$, $AP^{c}$, and $AP^{f}$ for rare, common, frequent categories on LVIS dataset.
Detailed configurations and hyper-parameter settings can be found in the supplementary.

\paragraph{ADE20K and CityScapes.}
Following~\cite{he2022masked}, we conduct semantic segmentation experiments on ADE20K~\cite{zhou2017scene} and CityScapes~\cite{cordts2016cityscapes} using UPerNet~\cite{xiao2018unified}.
We set the batch size as 16/8, distributed on 8 GPUs, and use random cropping with the size of 512/1024, random flipping with a probability of 0.5, and random photometric distortion for ADE20k/CityScapes when training.
The training length is set as 160K iterations with 1500 iterations for warming up, and we use AdamW optimizer with the learning rate of 1e-4, $\beta_1$ of 0.9, $\beta_2$ of 0.999, and weight decay of 0.05.
Poly learning rate decay and layer-wise learning rate decay are adopted.
Please refer to the supplementary for other configurations and hyper-parameter settings.

\begin{table}[t]
  \caption{Results on different downstream tasks using different target encoders. ``RGB'' denotes regressing the original RGB statistics as in the original MAE~\cite{he2022masked}.}\label{tab:recons_tar}
  \tablestyle{3.6pt}{1.5}
  \begin{tabular}{c|c|ccccc}
  \shline \hline
  \multirow{2}{*}{Reconstruction Target} & \multirow{2}{*}{Pre-train Dataset} & IN1K & \multicolumn{2}{c}{COCO} & ADE20K & Cityscapes \\ \cline{3-7}
   &  & Top-1 Acc. & $AP^{box}$ & $AP^{mask}$ & mIoU & mIoU \\ \shline
  \multirow{2}{*}{RGB~\cite{he2022masked}} & IN1K  & 82.90 & 50.27 & 44.95 & 44.17 & 80.44 \\ \cline{2-7} 
                               & IN22K & 82.74 & 51.08 & 45.54 & 45.21 & 80.15 \\ \shline
  \multirow{2}{*}{MAE~\cite{he2022masked}} & IN1K  & 83.31 & 51.83 & 45.99 & 47.20 & 81.59 \\ \cline{2-7} 
                               & IN22K & 83.56 & 52.23 & 46.33 & 47.80 & 81.05 \\ \shline
  \multirow{2}{*}{DINO~\cite{caron2021emerging}} & IN1K & 83.79 & 51.08 & 45.31 & 47.75 & 80.52 \\ \cline{2-7} 
                               & IN22K & 83.99 & 51.10 & 45.25 & 48.54 & 80.67 \\ \shline
  \multirow{2}{*}{CMAE~\cite{huang2022contrastive}} & IN1K & 83.82 & 52.27 & 46.46 & 50.05 & 82.19 \\ \cline{2-7} 
                               & IN22K & 83.84 & 52.83 & 47.02 & 49.56 & 82.13 \\ \shline
  \multirow{2}{*}{CLIP~\cite{radford2021learning}} & IN1K & 84.40 & 50.58 & 44.81 & 50.89 & 80.47 \\ \cline{2-7} 
                               & IN22K & 84.77 & 51.33 & 45.48 & 51.87 & 81.72 \\ \shline \hline
  \end{tabular}
  \end{table}

\section{Observations}
\subsection{Different Reconstruction Targets}
We first investigate the influence brought by different reconstruction targets using ImageNet-1k~\cite{deng2009imagenet} and ImageNet-22k~\cite{russakovsky2015imagenet} as pre-training datasets.
Self-supervised methods like MAE~\cite{he2022masked}, DINO~\cite{caron2021emerging}, CMAE~\cite{huang2022contrastive} and language-assisted methods like CLIP~\cite{radford2021learning} are selected as the target encoder to produce reconstruction targets.
The original MAE, which uses RGB images as the target, is adopted as baselines.
Results are shown in~\tabref{tab:recons_tar}.

Different reconstruction targets bring different semantic signals for models.
Directly regressing RGB statistics is proved to be effective in many recent works~\cite{he2022masked,xie2022simmim}.
However, compared with other reconstruction targets, simply using the image as the reconstruction target performs the worst on various downstream tasks due to the low-level statistics it learns~\cite{wei2022mvp,hou2022milan}.
We select four different target encoders: 
MAE~\cite{he2022masked} which is pre-trained in a pure MIM way; 
DINO~\cite{caron2021emerging} which is pre-trained in a pure augmentation-based contrastive learning way;
CMAE~\cite{huang2022contrastive} which is pre-trained under the combination of MIM and contrastive learning; 
CLIP~\cite{radford2021learning} which is pre-trained with language assistance.

When the dataset changes from ImageNet-1k to ImageNet-22k, which resulted in a significant increase in the number of images from \textasciitilde 1M to \textasciitilde 14M, we observe a noticeable improvement in the performance on the downstream tasks, especially on dense prediction tasks.
For instance, when CMAE is adopted as the target encoder on ImageNet-22k, we can achieve \textasciitilde  0.6\% and \textasciitilde 0.6\% performance gain with respect to $AP^{box}$ and $AP^{mask}$ compared with the model pre-trained on ImageNet-1k.
Under most scenarios, the model pre-trained with CLIP~\cite{radford2021learning} produces better results.
For example, on ImageNet-1k, the model can achieve 84.77\% Top-1 Accuracy when pre-trained on ImageNet-22k dataset, 1.87\% higher than the baseline (84.77\% vs. 82.90\%).
In addition, the model using CLIP as the target encoder in pre-training shows better data scaling ability.
Specifically, with CLIP's assistance, the performance increases on all four downstream tasks, 84.40\% $\rightarrow$ 84.77\% (+0.4\%) on ImageNet-1k, 80.47\% $\rightarrow$ 81.72\% (+1.25\%) on CityScapes~\etc
Above experimental results demonstrate that \textit{CLIP is a strong target encoder for pre-training of masked image modeling}, so we select CLIP as the target encoder for the following experiments.

\begin{figure}[t]
  \centering
  \includegraphics[width=0.98\textwidth]{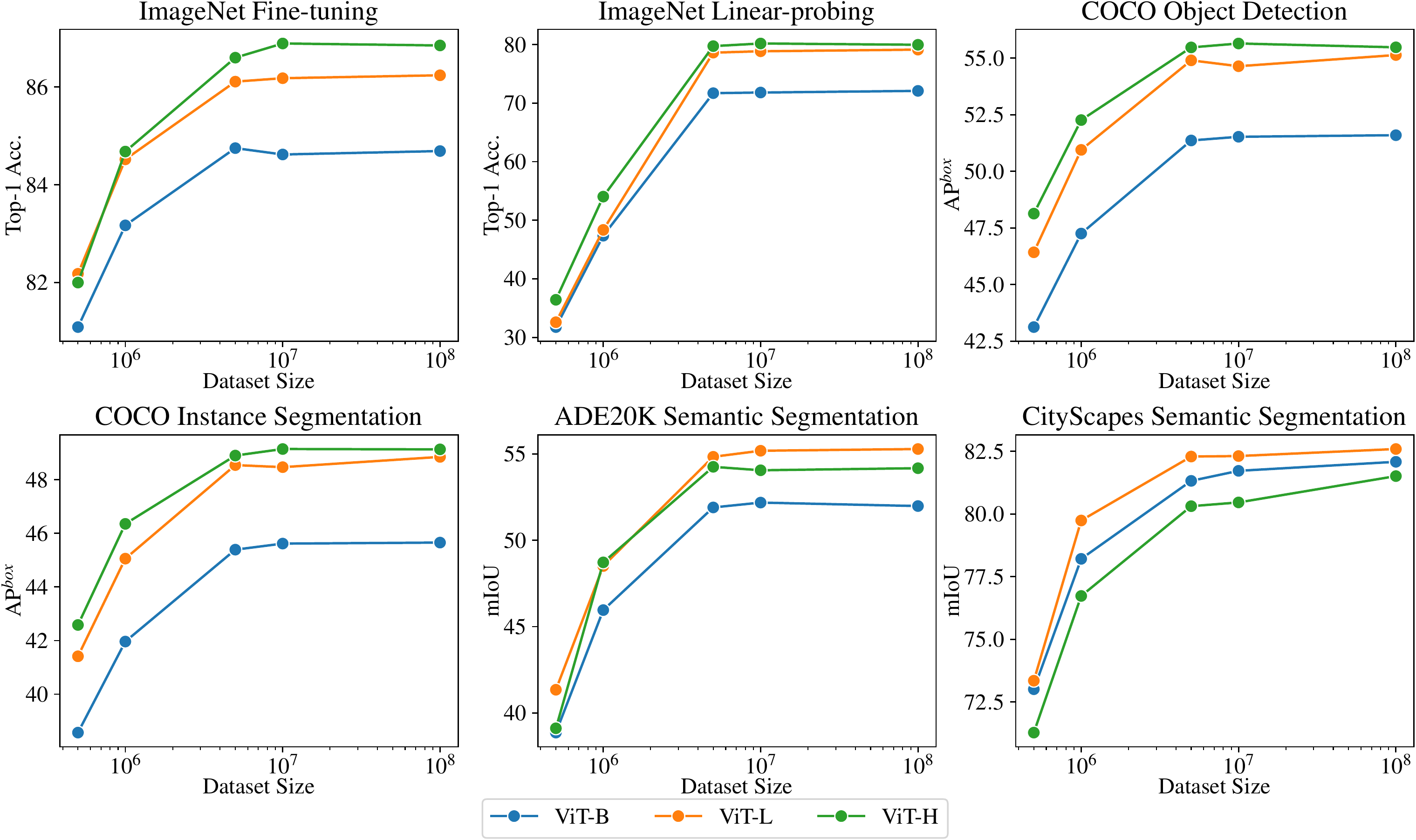}
  \caption{Visualization of relation curves between fine-tuning performances and different sizes of pre-training datasets. Various downstream tasks are evaluated including ImageNet-1k classification, COCO object detection, ADE20K semantic segmentation, and CityScapes semantic segmentation.}
  \label{fig:results_diff_dataset_size}
\end{figure}

\begin{table}[t]
  \tablestyle{6pt}{1.5}
  \caption{Fine-tuning and linear-probing results on ImageNet-1k~\cite{deng2009imagenet}. Models are pre-trained on  datasets with different sizes randomly sampled from Coyo-700M.}\label{tab:im1k_res}
  \centering
  \begin{tabular}{c|cccccccccc}
  \shline \hline
   \multirow{2}{*}{Model} & \multicolumn{2}{c}{Coyo-0.5M} & \multicolumn{2}{c}{Coyo-1M} & \multicolumn{2}{c}{Coyo-5M} & \multicolumn{2}{c}{Coyo-10M} & \multicolumn{2}{c}{Coyo-100M} \\ \cline{2-11} 
   & FT & LP & FT & LP & FT & LP & FT & LP & FT & LP \\ \hline
  ViT-B & 81.09 & 31.77 & 83.17 & 47.37 & 84.75 & 71.70 & 84.62 & 71.81 & 84.69 & 72.09 \\ \hline
  ViT-L & 82.18 & 32.60 & 84.52 & 48.36 & 86.11 & 78.62 & 86.18 & 78.85 & 86.24 & 79.13 \\ \hline
  ViT-H & 82.00 & 36.43 & 84.68 & 54.04 & 86.60 & 79.73 & 86.89 & 80.18 & 86.85 & 79.96 \\ \shline \hline
  \end{tabular}
  \end{table}

  \begin{table}[t]
    \tablestyle{2.0pt}{1.5}
    \caption{Results of object detection and instance segmentation on Microsoft COCO dataset,~\wrt different sizes of pre-training datasets.}\label{tab:det_res}
    \begin{tabular}{c|cccccccccc}
    \shline \hline
    \multirow{2}{*}{Model} & \multicolumn{2}{c}{Coyo-0.5M} & \multicolumn{2}{c}{Coyo-1M} & \multicolumn{2}{c}{Coyo-5M} & \multicolumn{2}{c}{Coyo-10M} & \multicolumn{2}{c}{Coyo-100M} \\ \cline{2-11} 
     & $AP^{box}$ & $AP^{mask}$ & $AP^{box}$ & $AP^{mask}$ & $AP^{box}$ & $AP^{mask}$ & $AP^{box}$ & $AP^{mask}$ & $AP^{box}$ & $AP^{mask}$ \\ \hline
    ViT-B & 43.12 & 38.57 & 47.25 & 41.96 & 51.37 & 45.39 & 51.53 & 45.62 & 51.60 & 45.65 \\ \hline
    ViT-L & 46.43 & 41.41 & 50.95 & 45.06 & 54.90 & 48.55 & 54.64 & 48.47 & 55.14 & 48.85 \\ \hline
    ViT-H & 48.13 & 42.58 & 52.27 & 46.35 & 55.48 & 48.90 & 55.65 & 49.15 & 55.48 & 49.13 \\ \shline \hline
    \end{tabular}
  \end{table}

\subsection{More Diverse Coyo Dataset}\label{sec:coyo_dataset}
With the rapid development of the multi-modal community, more and more web-scaled datasets, such as LAION-400M~\cite{schuhmann2021laion} and Coyo-700M~\cite{kakaobrain2022coyo-700m}, have been made publicly available, facilitating their accessibility for the wider community.
Meanwhile, the ImageNet-1k dataset~\cite{deng2009imagenet} is object-centric, which is inconsistent with real-world scenarios.
Hence, we choose Coyo-700M, which contains large-scale informative image-text pairs, for further study.
We visualize the results on different downstream tasks in~\figref{fig:results_diff_dataset_size}, and the quantitative results are shown in~\tabref{tab:im1k_res},~\tabref{tab:det_res}, and~\tabref{tab:seg_res}.

From~\figref{fig:results_diff_dataset_size}, we can easily observe that \textit{fine-tuning performances saturate on all the five downstream tasks, even when using ViT-H model with 100M images for pre-training}.
When the size of the pre-training dataset increases from 0.5M to 1M, the performance of the model improves abruptly. 
When the size of the pre-trained dataset is within the range of 1M to 10M, the  performance can still be sustainably improved in most cases, but the sign of performance saturation begins to appear.
Finally, when the size of the pre-trained dataset reaches 100M, performance on most downstream tasks hardly improve, and in some cases there is even a degradation in performance.
\textit{We conclude that when the size of the pre-trained dataset is limited to 10M, the model has strong data scalability. 
However, as the size of the pre-training dataset continues to increase, MIM-based pre-training is difficult to provide scalability for the model.}

Additionally, for ViT-H model, we observe that it sometimes produces worse results than ViT-L model, especially when pre-trained with a small dataset.
For example, when pre-trained using 0.5M images, ViT-H achieves 82.00\% Top-1 Accuracy on ImageNet-1k, which is nearly 0.2\% lower than ViT-L model.
We speculate that huge models still demand large-scale pre-training datasets to achieve better performances, but still, they cannot break the limit of performance saturation.

We also adopt linear-probing on ImageNet-1k to evaluate whether the model could scale up with a larger pre-training dataset.
Linear-probing freezes the backbone network and only tunes the classification head part.
Although it is not correlated with the transfer learning performance~\cite{he2022masked}, linear-probing is still a popular evaluation metric for representation learning.
Results of linear-probing are shown in~\tabref{tab:im1k_res}.
Under linear-probing, the scale of the pre-training dataset plays an important role when the domain of pre-training data differs from the validation set.
When the size of pre-training data is small, there exists a gap between the learned representation and validation set, which leads to poor performance (\eg 36.43\% achieved by ViT-H with 0.5M for pre-training).
With the size scales up to 5M, the performance of linear-probing increases sharply, achieving more than 20\% accuracy gain.
Nonetheless, MIM pre-training shows limited performance improvement when the pre-training data reach 100M.

\begin{table}[t]
  \tablestyle{6pt}{1.5}
  \caption{Results of semantic segmentation on ADE20K~\cite{zhou2017scene} and CityScapes~\cite{cordts2016cityscapes},~\wrt different sizes of pre-training datasets.}\label{tab:seg_res}
  \begin{tabular}{c|cccccccccc}
  \shline \hline
  \multirow{2}{*}{Model} & \multicolumn{2}{c}{Coyo-0.5M} & \multicolumn{2}{c}{Coyo-1M} & \multicolumn{2}{c}{Coyo-5M} & \multicolumn{2}{c}{Coyo-10M} & \multicolumn{2}{c}{Coyo-100M} \\ \cline{2-11} 
   & ADE. & City. & ADE. & City. & ADE. & City. & ADE. & City. & ADE. & City. \\ \hline
  ViT-B & 38.87 & 73.01 & 45.96 & 78.21 & 51.90 & 81.32 & 52.18 & 81.72 & 51.98 & 82.08 \\ \hline
  ViT-L & 41.35 & 73.35 & 48.51 & 79.74 & 54.83 & 82.29 & 55.18 & 82.31 & 55.28 & 82.59 \\ \hline
  ViT-H & 39.12 & 71.28 & 48.72 & 76.73 & 54.25 & 80.31 & 54.05 & 80.46 & 54.17 & 81.51 \\ \shline \hline
  \end{tabular}
  \vspace{-2mm}
  \end{table}

\begin{table}[t]
\centering
\caption{We fine-tune ViT-L/16 and report the results of different data sampling strategies and results are shown in a ``A/B'' format, where ``A'' denotes randomly sampling and ``B'' denotes the strategy proposed in CiT~\cite{xu2023cit}.}\label{tab:data_sampling_strategy}
  \tablestyle{12pt}{1.5}
  \begin{tabular}{cccc}
  \shline \hline
  Dataset  & Coyo-1M & Coyo-10M & Coyo-100M \\
  \hline
  ImageNet-1k & 84.52 / 84.59 & 86.18 / 86.20 & 86.24 / 86.22 \\
  COCO & 50.95 / 50.59 & 54.64 / 54.89 & 55.14 / 54.99 \\
  ADE20K & 48.51 / 49.84 & 55.18 / 55.34 & 55.28 / 55.56 \\
  CityScapes & 79.74 / 78.48 & 82.31 / 82.80 & 82.59 / 82.59 \\
  \shline \hline
\end{tabular}
\end{table}

\subsection{Better Data Sampling Strategy}
Noticing that pre-training using Coyo-1M finally achieves 83.17\% Top-1 Accuracy on ImageNet-1k in~\tabref{tab:im1k_res}, which is much lower than the one pre-trained with ImageNet-1k (83.17\%~ vs. 84.40\%), 
we conjecture that using data from the same domain for pre-training and validation results in the performance gap, and the quality of pre-training data plays an important role.
To evaluate whether the data scaling ability is restricted by the quality of pre-training data, we use the sampling strategy proposed in CiT~\cite{xu2023cit} instead of randomly sampling to select images from Coyo-700M for pre-training.
CiT measures the similarity of text embeddings between the metadata and the raw data and selects training data relevant to tasks of interest in an online way.
Here, we adopt the text encoder from EVA~\cite{fang2022eva} and compare the similarity between the text description provided by Coyo-700M dataset and the class labels from ImageNet-1k.
We set different threshold values to sample the pre-training dataset with different scales in an offline way.
As shown in~\tabref{tab:data_sampling_strategy}, CiT sampling strategy does not lead to any performance improvement.
We hypothesize that \textit{MIM pre-training is data agnostic, which means that whether pre-training data is simple or complex does not influence the performance, only if we add pre-training data from the same domain with the validation set.}
Meanwhile, a ``better'' data sampling strategy does not change the tendency of data scaling.
With the sampling strategy of CiT, ViT-L can rapidly obtain the performance gain on ImageNet-1k from Coyo-1M to Coyo-10M (\ie 84.59\% $\rightarrow$ 86.20\%), while the performance nearly freezes when (\ie 86.20\% $\rightarrow$ 86.22\%) the size grows to 100M.

\subsection{``Harder'' Downstream Tasks}
In order to explore whether the downstream tasks limit the capacity of large-scale pre-trained models, we try to build or evaluate masked image modeling on ``harder'' downstream tasks,
including classification on ImageNet-5k, long-tailed classification on iNaturalist2018~\cite{van2018inaturalist}, long-tailed object detection on LVIS~\cite{gupta2019lvis}.
Results are shown in~\tabref{tab:hard_downstream_strong_encoder} and~\tabref{tab:lvis}.

From~\tabref{tab:hard_downstream_strong_encoder}, on ImageNet-5k, which contains more classes than ImageNet-1k, we can easily find that
the fine-tuning performance increases quickly when the data size scales up to 10M (59.09\% vs. 58.01\%).
However, it is still difficult for the model to scale up to 100M pre-training data, which only achieves \textasciitilde 0.3\% performance gain (59.09\% vs. 59.38\%).
This phenomenon is consistent with the trends we have observed when fine-tuning on ImageNet-1k.
In addition, we also evaluate ViT-L/16 on iNaturalist2018 fine-grained image classification, and observe a similar pattern.
Note that in~\tabref{tab:hard_downstream_strong_encoder}, the performance of using Coyo-5M for pre-training even surpasses the one of using Coyo-10M (81.09\% vs. 80.61\%).
The main reason, we think, is that the model pre-trained by Coyo-5M is trained for much more iterations.

We also report the results on LVIS~\cite{gupta2019lvis}, which is a challenging dataset with large-vocabulary object categories as well as high-quality instance masks.
Unlike COCO~\cite{lin2014microsoft}, LVIS contains a large number of rare categories, and we adopt the metric defined in~\cite{gupta2019lvis} for better evaluating the ability to detect long-tailed classes.
From~\tabref{tab:lvis}, we observe that with the larger dataset for pre-training, the model can achieve better performances significantly.
Specifically, ViT-L achieves the best $AP_{box}$ of 47.30\% with Coyo-100M for pre-training, better than the performance of the model pre-trained with Coyo-10M by 0.98\% (47.30\% vs. 46.32\%).
Furthermore, we find that the performance gain mainly comes from the rare category.
When the size of the dataset increases from 10M to 100M, the performance of rare classes boosts over 3\% (36.51\% $\rightarrow$ 39.83\%), while the performance of frequent classes only increases less than 1\% (50.93\% $\rightarrow$ 51.59\%), indicating that large-scale data pre-training may help in long-tailed object detection, as well as instance segmentation.

We also test the data scalability on various robustness datasets, including ImageNet-A~\cite{hendrycks2021natural}, ImageNet-R~\cite{hendrycks2021many}, ImageNet-C~\cite{hendrycks2019benchmarking}, COCO-C~\cite{hendrycks2019benchmarking}, and CityScapes-C~\cite{hendrycks2019benchmarking}.
The conclusion is nearly the same as the above, and detailed test results can be found in the supplementary materials.

\begin{table}[t]
  \centering
  \caption{We report the results of fine-tuning on ``harder'' downstream tasks or using stronger target encoders.}\label{tab:hard_downstream_strong_encoder}
  \tablestyle{1.6pt}{1.5}
  \begin{tabular}{cccccccc}
  \shline \hline
  Dataset & Model & Target Encoder & Coyo-0.5M & Coyo-1M & Coyo-5M & Coyo-10M & Coyo-100M \\
  \hline
  iNat2018 & ViT-L/16 & CLIP-B/16 & 67.8 & 74.51 & 81.09 & 80.61 & 81.28 \\ 
  ImageNet-5k & ViT-L/16 & CLIP-B/16 & - & 58.01 & - & 59.09 & 59.38 \\
  \hline
  ImageNet-1k & ViT-L/16 & CLIP-B/16  & 82.18 & 84.52 & 86.11 & 86.18 & 86.24 \\
  ImageNet-1k & ViT-L/14 & CLIP-L/14  &   -   &   -   &   -   &   -   & 86.66 \\
  ImageNet-1k & $\text{ViT-L/14}^{\dag}$ & EVA-G/14   &   -   &   -   &   -   &   -   & 86.90 \\
  ImageNet-1k & ViT-L/14 & EVA-G/14   & 82.66 & 85.00 & 86.91 & 86.93 & 86.94 \\
  \shline \hline
\end{tabular}
\end{table}

\begin{table}[t]
\centering
\caption{Quantitative results of ViT-L/16 evaluated on LVIS validation set~\cite{gupta2019lvis}. ``r'', ``c'', and ``f'' represent ``rare'', ``common'', and ``frequent'' respectively.
MIM pre-training infrequently shows strong data scalability on rare categories.}~\label{tab:lvis}
\tablestyle{3pt}{1.5}
\begin{tabular}{c|cccccccc}
\shline \hline
Pre-training Dataset & $AP^{box}$ & $AP^{box}_r$ & $AP^{box}_c$ & $AP^{box}_f$ & $AP^{mask}$ & $AP^{mask}_r$ & $AP^{mask}_c$ & $AP^{mask}_f$ \\ \hline
Coyo-0.5M & 28.00 & 21.44 & 31.28 & 39.37 & 27.28 & 22.20 & 30.29 & 36.24 \\ 
Coyo-1M   & 39.07 & 29.01 & 37.65 & 45.09 & 37.16 & 29.75 & 36.45 & 41.22 \\ 
Coyo-5M   & 46.32 & 35.80 & 45.93 & 51.37 & 43.26 & 34.88 & 43.48 & 46.70 \\ 
Coyo-10M  & 46.32 & 36.51 & 46.05 & 50.93 & 43.53 & 35.93 & 43.84 & 46.53 \\ 
Coyo-100M & 47.30 & 39.83 & 46.41 & 51.59 & 44.33 & 38.82 & 44.05 & 47.08 \\ 
\shline \hline

\end{tabular}
\end{table}

\subsection{Stronger Target Encoder}
Using ViT-B/16 as the target encoder is a compromise on the training cost, which may limit the data scalability of the model.
We therefore also try much larger target encoders for reconstruction, including ViT-L/14 (\textasciitilde 650M parameters) from CLIP~\cite{radford2021learning}, and EVA-G/14 (\textasciitilde 1B parameters) from EVA~\cite{fang2022eva}.
The results are listed in~\tabref{tab:hard_downstream_strong_encoder}.

When pre-trained with 100M data, using CLIP-L/14 achieves 86.66\% Top-1 Accuracy, which is about 0.4\% higher than the model pre-trained with CLIP-B/16.
However, it may be caused by the difference in patch size.
Here, we use a stronger target encoder, EVA-G/14 with about 1.0B parameters, for reconstruction and it achieves 86.94\% Top-1 Accuracy, about 0.3\% higher than the model pre-trained with CLIP-L/14 (86.94\% vs. 86.66\%).
We therefore believe that \textit{using a stronger target encoder may help in increasing the capacity of the model.}
Then, to investigate whether longer pre-training helps in the performance, we pre-train an encoder with 10 epochs using 100M data. The model reaches 86.90\%, which is similar to the model pre-trained with 15 epochs (86.90\% vs. 86.94\%).
In other words, \textit{when the performance tends to saturate, it is hard to improve the performance even with longer pre-training epochs.}
Last, we use EVA-G/14, which contains \textasciitilde1B parameters, as the target encoder to explore whether the encoder pre-trained in a MIM manner is scalable on data.
Unfortunately, we obtain the same conclusion that the model pre-trained with masked image modeling is hard to scale on more pre-training data when the model size is fixed.

\section{Conclusion}
In this study, we delve deeper into the data scaling capabilities of masked image modeling.
Unlike previous work~\cite{xie2022data}, we undertake MIM pre-training using the extensive web-scale Coyo-700M dataset~\cite{kakaobrain2022coyo-700m} and observe that the data scalability of masked image modeling is roughly limited to 10M data. 
When we conduct pre-training on a larger dataset, masked image modeling struggles to learn improved local relations of images under most scenarios.
Despite conducting extensive experiments, including tackling more challenging downstream tasks and employing stronger target encoders, 
we observe limited performance gains in MIM pre-training, with the exception of experiments conducted on the LVIS~\cite{gupta2019lvis} dataset. 
These observations highlight a prevalent issue of performance saturation in MIM pre-training when scaling up to larger datasets. 
The challenge of achieving substantial performance improvements under large-scale pre-training remains unresolved for masked image modeling.
Continued research and innovation to address the performance saturation challenge of masked image modeling are especially needed to unlock the full potential of masked image modeling in the context of large-scale pre-training.

\paragraph{Limitations.}
First, we only provide experimental observations without solutions to data scaling problems for masked image modeling.
How to solve this tough problem still needs to be explored. 
Second, the detailed configurations are suboptimal.
We do not search for better training recipes for pre-training or fine-tuning due to the complexity of hyper-parameters.
Third, we only adopt the MAE-style~\cite{he2022masked} pre-training and fine-tuning to represent MIM methods.
Many recent works~\cite{gao2022convmae,liu2022devil,bao2021beit} are not involved.
Fourth, the scale of models and datasets is still limited and relatively small compared with recent multi-modal models.

{
\small
\bibliographystyle{IEEEtran}
\bibliography{mybib}
}

\appendix

\newpage

\section{Robustness Test}\label{sec:robust}
We report the performances on various robustness datasets, including ImageNet-A~\cite{hendrycks2021natural}, ImageNet-R~\cite{hendrycks2021many}, ImageNet-C~\cite{hendrycks2019benchmarking}, COCO-C~\cite{hendrycks2019benchmarking}, and CityScapes-C~\cite{hendrycks2019benchmarking}.
Here we first evaluate different reconstruction targets in~\tabref{tab:rec_tar_img_c}.
On robustness datasets, CLIP~\cite{radford2021learning}, as the target encoder, also performs the best. 
Then we select CLIP as the target encoder and use Coyo dataset~\cite{kakaobrain2022coyo-700m} for pre-training, and results are shown in~\tabref{tab:results_img_c},~\tabref{tab:results_coco_c}, and~\tabref{tab:results_city_c}.
Similar trends of data scaling can be found here.
The ability of data scaling is roughly limited to 10M.
When the scale exceeds 10M, it is hard for masked image modeling to boost the performances on robustness datasets, like ImageNet-C.
More conclusions can be found in~\secref{sec:coyo_dataset}.

\begin{table}[!h]
\caption{Tabular results on ImageNet-A/R/C with different reconstruction targets.}\label{tab:rec_tar_img_c}
\tablestyle{4pt}{1.5}
\begin{tabular}{c|c|ccccc}
\shline \hline
\multirow{2}{*}{Reconstruction Target} & \multirow{2}{*}{Pre-training Dataset} & \multirow{2}{*}{Clean} & \multirow{2}{*}{ImageNet-A} & \multirow{2}{*}{ImageNet-R} & \multicolumn{2}{c}{ImageNet-C} \\ \cline{6-7} 
 &  &  &  &  & \multicolumn{1}{l}{Acc.} & mCE($\downarrow$) \\ \shline
\multirow{2}{*}{RGB~\cite{he2022masked}} & ImageNet-1k & 82.9 & - & - & - & - \\ \cline{2-7} 
 & ImageNet-22k & 83.0 & 33.5 & 48.8 & 76.8 & 51.3 \\ \shline
\multirow{2}{*}{MAE~\cite{he2022masked}} & ImageNet-1k & 83.3 & 35.7 & 48.9 & 77.3 & 50.1 \\ \cline{2-7} 
 & ImageNet-22k & 83.6 & 37.4 & 49.8 & 77.5 & 49.5 \\ \shline
\multirow{2}{*}{DINO~\cite{caron2021emerging}} & ImageNet-1k & 83.8 & 40.3 & 50.4 & 78.5 & 46.8 \\ \cline{2-7} 
 & ImageNet-22k & 84.0 & 40.7 & 50.0 & 78.6 & 46.6 \\ \shline
\multirow{2}{*}{CMAE~\cite{huang2022contrastive}} & ImageNet-1k & 83.8 & 41.0 & 50.6 & 78.3 & 47.4 \\ \cline{2-7} 
 & ImageNet-22k & 83.9 & 41.9 & 51.0 & 78.3 & 46.3 \\ \shline
\multirow{2}{*}{CLIP~\cite{radford2021learning}} & ImageNet-1k &  84.4 & 44.3 & 51.3 & 78.9 & 46.3 \\ \cline{2-7} 
 & ImageNet-22k & 84.8 & 48.2 & 56.8 & 79.5 & 45.1 \\ \shline \hline
\end{tabular}
\end{table}

\newpage

\begin{table}[!h]
\caption{Tabular results on ImageNet-A/R/C with Coyo dataset~\cite{kakaobrain2022coyo-700m} for pre-training.}\label{tab:results_img_c}
\tablestyle{8pt}{1.1}
\begin{tabular}{c|c|ccccc}
\shline \hline
\multirow{2}{*}{Model} & \multirow{2}{*}{Pre-training Dataset} & \multirow{2}{*}{Clean} & \multirow{2}{*}{ImageNet-A} & \multirow{2}{*}{ImageNet-R} & \multicolumn{2}{c}{ImageNet-C} \\ \cline{6-7} 
 &  &  &  &  & Acc. & mCE \\ \shline \hline
\multirow{5}{*}{ViT-B} & Coyo-0.5M & 81.1 & 22.1 & 46.5 & 75.5 & 53.7 \\ \cline{2-7} 
 & Coyo-1M & 83.2 & 22.1 & 50.1 & 77.8 & 49.6 \\ \cline{2-7} 
 & Coyo-5M & 84.8 & 45.7 & 55.5 & 79.2 & 45.4 \\ \cline{2-7} 
 & Coyo-10M & 84.6 & 44.8 & 56.0 & 79.2 & 45.5 \\ \cline{2-7} 
 & Coyo-100M & 84.7 & 45.3 & 55.4 & 79.2 & 45.2 \\ \shline\hline
\multirow{5}{*}{ViT-L} & Coyo-0.5M & 82.2 & 29.7 & 51.1 & 77.3 & 49.5 \\ \cline{2-7} 
 & Coyo-1M & 84.5 & 46.0 & 57.7 & 79.7 & 44.1 \\ \cline{2-7} 
 & Coyo-5M & 86.1 & 61.6 & 67.2 & 81.6 & 37.8 \\ \cline{2-7} 
 & Coyo-10M & 86.2 & 62.1 & 67.6 & 81.7 & 37.3 \\ \cline{2-7} 
 & Coyo-100M & 86.2 & 62.4 & 67.8 & 81.7 & 37.4 \\ \shline \hline
\multirow{5}{*}{ViT-H} & Coyo-0.5M & 82.0 & 28.4 & 51.8 & 77.1 & 50.1 \\ \cline{2-7} 
 & Coyo-1M & 84.7 & 46.9 & 60.7 & 80.3 & 43.1 \\ \cline{2-7} 
 & Coyo-5M & 86.6 & 67.5 & 72.1 & 82.5 & 35.0 \\ \cline{2-7} 
 & Coyo-10M & 86.9 & 67.8 & 73.1 & 82.5 & 35.3 \\ \cline{2-7} 
 & Coyo-100M & 86.9 & 68.5 & 73.2 & 82.7 & 34.5 \\ \shline \hline
\end{tabular}
\end{table}

\begin{table}[!h]
\caption{Tabular results on COCO-C with Coyo dataset~\cite{kakaobrain2022coyo-700m} for pre-training.}\label{tab:results_coco_c}
\tablestyle{4pt}{1.5}
\resizebox{\linewidth}{!}{
\begin{tabular}{c|c|c|cccc|cccc|cccc|cccc}
\shline\hline
\multirow{2}{*}{Model} & Pre-training & \multirow{2}{*}{Average} & \multicolumn{4}{c}{Noise} & \multicolumn{4}{c}{Blur} & \multicolumn{4}{c}{Digital} & \multicolumn{4}{c}{Weather} \\ \cline{4-19} 
 &  Dataset & & Gauss & Shot & Speck & Impul & Gauss & Motion & Glass & Defoc & Contr & JPEG & Satur & Pixel & Snow & Fog & Frost & Spatt \\ \shline\hline
\multirow{5}{*}{ViT-B} & Coyo-0.5M  & 24.5 & 19.6 & 18.8 & 24.1 & 17.7 & 27.1 & 22.1 & 18.7 & 25.5 & 23.8 & 26.4 & 36.4 & 19.0 & 22.3 & 33.3 & 25.3 & 32.2  \\ \cline{2-19} 
 & Coyo-1M  & 29.3 & 24.4 & 23.9 & 28.9 & 22.9 & 30.3 & 25.0 & 21.4 & 28.1 & 30.4 & 30.7 & 41.3 & 26.9 & 27.6 & 39.9 & 30.2 & 37.3 \\ \cline{2-19} 
 & Coyo-5M & 36.1 & 31.7 & 31.5 & 36.7 & 31.3 & 34.4 & 30.2 & 27.2 & 32.0 & 37.9 & 37.5 & 47.0 & 37.5 & 35.6 & 46.1 & 37.6 & 44.0 \\ \cline{2-19} 
 & Coyo-10M & 36.7 & 32.7 & 32.5 & 37.5 & 32.1 & 34.7 & 30.4 & 26.6 & 32.2 & 39.6 & 37.2 & 47.4 & 38.2 & 36.1 & 47.0 & 38.1 & 44.7 \\ \cline{2-19} 
 & Coyo-100M & 36.5 & 32.2 & 32.0 & 36.9 & 32.0 & 34.7 & 30.1 & 26.5 & 32.2 & 39.7 & 37.2 & 47.2 & 37.2 & 36.4 & 47.0 & 38.0 & 44.8 \\ \shline\hline
\multirow{5}{*}{ViT-L} & Coyo-0.5M  & 30.1 & 25.9 & 25.4 & 30.3 & 24.9 & 30.8 & 25.3 & 22.3 & 28.6 & 32.3 & 31.4 & 40.8 & 24.8 & 28.9 & 40.0 & 31.4 & 37.9  \\ \cline{2-19} 
 & Coyo-1M  & 35.9 & 32.1 & 31.6 & 36.3 & 31.5 & 34.6 & 29.8 & 25.8 & 32.3 & 39.2 & 36.3 & 46.3 & 35.2 & 35.7 & 46.4 & 37.3 & 43.7 \\ \cline{2-19} 
 & Coyo-5M & 41.8 & 39.0 & 39.1 & 44.4 & 39.5 & 38.1 & 33.7 & 30.7 & 35.5 & 46.0 & 40.9 & 51.1 & 43.9 & 43.2 & 50.9 & 44.0 & 48.8 \\ \cline{2-19} 
 & Coyo-10M  & 42.3 & 39.5 & 39.9 & 44.3 & 39.7 & 38.6 & 34.0 & 31.2 & 36.0 & 46.1 & 41.9 & 51.5 & 44.4 & 43.3 & 51.7 & 44.1 & 49.8 \\ \cline{2-19} 
 & Coyo-100M  & 42.8 & 40.2 & 40.5 & 44.7 & 40.5 & 38.7 & 34.3 & 32.1 & 35.9 & 47.1 & 43.4 & 51.8 & 44.4 & 43.7 & 52.2 & 44.5 & 50.2 \\ \shline\hline
\multirow{5}{*}{ViT-H} & Coyo-0.5M & 32.5 & 28.4 & 27.9 & 33.0 & 27.9 & 33.0 & 27.2 & 24.4 & 30.9 & 34.3 & 33.9 & 42.9 & 28.7 & 31.4 & 42.1 & 33.6 & 39.9 \\ \cline{2-19} 
 & Coyo-1M  & 38.0 & 34.7 & 34.5 & 39.3 & 34.5 & 36.2 & 31.5 & 27.6 & 33.5 & 41.5 & 39.3 & 47.9 & 38.5 & 37.2 & 47.8 & 38.7 & 45.5 \\ \cline{2-19} 
 & Coyo-5M & 43.3 & 41.0 & 41.1 & 45.4 & 41.6 & 39.4 & 34.6 & 32.6 & 36.5 & 46.5 & 44.8 & 52.2 & 44.7 & 44.1 & 52.4 & 44.8 & 50.9 \\ \cline{2-19} 
 & Coyo-10M & 43.7 & 41.2 & 41.5 & 45.7 & 41.6 & 39.8 & 35.4 & 33.3 & 37.1 & 47.1 & 44.6 & 52.6 & 45.3 & 44.8 & 52.6 & 45.2 & 51.3 \\ \cline{2-19} 
 & Coyo-100M & 43.6 & 41.3 & 41.5 & 45.6 & 41.8 & 39.5 & 34.8 & 32.8 & 36.7 & 47.6 & 44.8 & 52.4 & 44.7 & 44.7 & 52.6 & 45.1 & 50.9 \\ \shline\hline
\end{tabular}
}
\end{table}

\begin{table}[!h]
\caption{Tabular results on CityScapes-C with Coyo dataset~\cite{kakaobrain2022coyo-700m} for pre-training.}\label{tab:results_city_c}
\tablestyle{4pt}{1.5}
\resizebox{\linewidth}{!}{
\begin{tabular}{c|c|c|cccc|cccc|cccc|cccc}
\shline\hline
\multirow{2}{*}{Model} & Pre-training  & \multirow{2}{*}{Average} & \multicolumn{4}{c}{Noise} & \multicolumn{4}{c}{Blur} & \multicolumn{4}{c}{Digital} & \multicolumn{4}{c}{Weather} \\ \cline{4-19} 
 &  Dataset &   & Gauss & Shot & Speck & Impul & Gauss & Motion & Glass & Defoc & Contr & JPEG & Satur & Pixel & Snow & Fog & Frost & Spatt \\ \shline\hline
\multirow{5}{*}{ViT-B} & Coyo-0.5M & 46.0 & 14.8 & 18.4 & 36.1 & 11.4 & 65.8 & 64.7 & 55.4 & 63.5 & 66.7 & 36.9 & 71.4 & 58.6 & 26.2 & 60.5 & 32.5 & 54.0 \\ \cline{2-19} 
 & Coyo-1M & 56.8 & 32.9 & 37.5 & 53.8 & 29.7 & 71.5 & 69.3 & 61.5 & 69.5 & 74.3 & 51.4 & 76.9 & 69.0 & 33.7 & 69.5 & 44.8 & 63.3 \\ \cline{2-19} 
 & Coyo-5M & 67.3 & 48.6 & 54.3 & 66.5 & 45.6 & 75.6 & 74.1 & 68.9 & 73.8 & 79.4 & 63.7 & 80.7 & 77.7 & 58.9 & 76.7 & 59.0 & 73.5 \\ \cline{2-19} 
 & Coyo-10M & 68.2 & 52.3 & 57.7 & 67.8 & 50.1 & 76.0 & 74.1 & 69.5 & 74.2 & 80.2 & 63.7 & 80.9 & 78.2 & 59.4 & 77.6 & 56.1 & 73.4 \\ \cline{2-19} 
 & Coyo-100M & 67.7 & 49.2 & 54.6 & 66.4 & 46.8 & 76.0 & 74.3 & 68.7 & 74.1 & 80.3 & 63.1 & 81.2 & 78.9 & 59.0 & 78.0 & 58.4 & 73.7 \\ \shline\hline
\multirow{5}{*}{ViT-L} & Coyo-0.5M & 51.6 & 27.1 & 30.5 & 46.1 & 21.7 & 66.2 & 64.5 & 57.6 & 64.0 & 68.4 & 48.9 & 71.7 & 65.0 & 33.0 & 62.7 & 39.4 & 59.1  \\ \cline{2-19} 
 & Coyo-1M & 62.0 & 40.1 & 44.6 & 59.1 & 35.6 & 73.7 & 71.5 & 65.7 & 71.7 & 76.6 & 61.1 & 78.7 & 74.9 & 45.0 & 74.4 & 51.2 & 68.7 \\ \cline{2-19} 
 & Coyo-5M & 71.9 & 58.4 & 64.2 & 73.5 & 55.8 & 78.0 & 75.8 & 73.1 & 76.5 & 80.6 & 67.7 & 81.5 & 80.1 & 66.2 & 79.4 & 63.6 & 76.1 \\ \cline{2-19} 
 & Coyo-10M & 71.8 & 60.5 & 65.4 & 72.9 & 58.6 & 77.5 & 75.7 & 72.9 & 75.9 & 80.3 & 66.0 & 81.7 & 79.4 & 65.9 & 78.9 & 62.1 & 75.8 \\ \cline{2-19} 
 & Coyo-100M & 72.6 & 61.2 & 66.7 & 74.3 & 58.1 & 78.1 & 75.9 & 72.3 & 76.5 & 80.9 & 67.9 & 81.8 & 80.5 & 67.0 & 80.0 & 63.0 & 77.0 \\ \shline\hline
\end{tabular}
}
\vspace{5mm}
\end{table}

\section{Pre-training Configurations}\label{sec:pre_train}

\begin{table}[h!]
\centering
\caption{Hyper-parameters for pre-training.}\label{tab:pt_setting}
\tablestyle{8pt}{1.05}
\begin{tabular}{l|c}
\shline \hline
config & value \\
\hline
dataset & Coyo-\{0.5, 1, 5, 10, 100\}M \\
epochs & \{300, 300, 300, 90, 15\} \\
warmup epochs~\cite{goyal2017accurate} & \{40, 40, 40, 40, 1\} \\
optimizer & AdamW \cite{loshchilov2017decoupled} \\
base learning rate & 1.5e-4 \\
weight decay & 0.05 \\
optimizer momentum & $\beta_1, \beta_2{=}0.9, 0.95$ \\
batch size & 4096 \\
learning rate schedule & cosine decay~\cite{loshchilovsgdr} \\
augmentation & RandomResizedCrop \\ \shline \hline
\end{tabular}
\end{table}

\section{Configurations for Image Classification Fine-tuning}\label{sec:im1k_cfg}
\begin{table}[!h]
\centering
\caption{Hyper-parameters for ImageNet-1k~\cite{deng2009imagenet} / iNaturalist~\cite{van2018inaturalist} fine-tuning.}\label{tab:ft_setting}
\tablestyle{8pt}{1.05}
\begin{tabular}{l|c}
\shline \hline
config & value \\
\hline
model & ViT-\{B, L, H\}/16 \\
epochs & \{100, 50, 50\} \\
warmup epochs~\cite{goyal2017accurate} & 5 \\
optimizer & AdamW~\cite{loshchilov2017decoupled} \\
base learning rate & 1e-3 \\
weight decay & 0.05 \\
optimizer momentum & $\beta_1, \beta_2{=}0.9, 0.999$ \\
layer-wise lr decay~\cite{bao2021beit} & \{0.75, 0.75, 0.65\} \\
batch size & 1024 \\
learning rate schedule & cosine decay~\cite{loshchilovsgdr} \\
augmentation & RandAug (9, 0.5)~\cite{cubuk2020randaugment} \\
label smoothing~\cite{szegedy2016rethinking} & 0.1 \\
mixup~\cite{zhangmixup} & 0.8 \\
cutmix~\cite{yun2019cutmix} & 1.0 \\
random erase~\cite{zhong2020random} & 0.25 \\
drop path~\cite{huang2016deep} & {0.1, 0.2, 0.3} \\ \shline \hline
\end{tabular}
\end{table}

\section{Configurations for Linear-probing}\label{sec:lp_cfg}
\begin{table}[h]
\centering
\caption{Hyper-parameters for linear-probing.}\label{tab:lp_setting}
\tablestyle{8pt}{1.05}
\begin{tabular}{l|c}
\shline \hline
config & value \\
\hline
model & ViT-\{B, L, H\}/16 \\
epochs & \{90, 50, 50\}\\
warmup epochs~\cite{goyal2017accurate} & 10 \\
optimizer & LARS~\cite{you2017large} \\
base learning rate & 0.1 \\
weight decay & 0 \\
batch size & 16384 \\
learning rate schedule & cosine decay \\
augmentation & RandomResizedCrop \\ \shline \hline
\end{tabular}
\end{table}

\section{Configurations for Semantic Segmentation}\label{sec:seg_cfg}
\begin{table}[h!]
\centering
\caption{Hyper-parameters for ADE20K~\cite{ade20k} / CityScapes~\cite{cordts2016cityscapes} semantic segmentation.}\label{tab:seg_setting}
\tablestyle{8pt}{1.05}
\begin{tabular}{l|cc}
\shline \hline
config & ADE20K & CityScapes \\
\hline
input resolution & 512$\times$512 & 1024$\times$1024 \\
batch size & 16 & 8 \\
fine-tuning iterations & \multicolumn{2}{c}{160K} \\
warmup iterations & \multicolumn{2}{c}{1.5K} \\
optimizer & \multicolumn{2}{c}{AdamW~\cite{loshchilov2017decoupled}} \\
learning rate & \multicolumn{2}{c}{1e-4} \\
learning rate schedule & \multicolumn{2}{c}{poly} \\
weight decay & \multicolumn{2}{c}{0.05} \\
optimizer momentum & \multicolumn{2}{c}{$\beta_1, \beta_2{=}0.9, 0.999$} \\ \hline
model & \multicolumn{2}{c}{ViT-\{B, L, H\}/16} \\
layer-wise lr decay~\cite{bao2021beit} & \multicolumn{2}{c}{0.65} \\
drop path~\cite{huang2016deep} & \multicolumn{2}{c}{\{0.1, 0.2, 0.3\} }\\
\shline \hline
\end{tabular}
\end{table}

\end{document}